\documentclass[10pt,twocolumn,letterpaper]{article}

\usepackage{cvpr}              

\usepackage{graphicx}
\usepackage{amsmath}
\usepackage{amssymb}
\usepackage{amsxtra}
\usepackage{booktabs}

\usepackage{mathrsfs}
\usepackage{amsthm}
\usepackage{multirow}
\usepackage{makecell}
\usepackage{dsfont}

\usepackage{float}

\usepackage[dvipsnames]{xcolor}

\definecolor{cvprblue}{rgb}{0.21,0.49,0.74}
\usepackage[pagebackref,breaklinks,colorlinks,citecolor=cvprblue]{hyperref}
\usepackage{marvosym}

\usepackage[capitalize]{cleveref}
\crefname{section}{Sec.}{Secs.}
\Crefname{section}{Section}{Sections}
\Crefname{table}{Table}{Tables}
\crefname{table}{Tab.}{Tabs.}

\def\paperID{1608} 
\def\confName{CVPR}
\def\confYear{2024}
\begin{document}

\title{OED: Towards One-stage End-to-End Dynamic Scene Graph Generation}


\author{Guan Wang$^1$ \quad Zhimin Li$^2$ \quad Qingchao Chen$^3$ \quad Yang Liu$^{1}$\thanks{Corresponding author}\\
$^1$Wangxuan Institute of Computer Technology, Peking University \\
$^2$Tencent Inc. \quad
$^3$National Institute of Health Data Science, Peking University \\
{\tt\small w.g@stu.pku.edu.cn} \quad 
{\tt\small zhiminli.cn@outlook.com} \quad
{\tt\small \{qingchao.chen, yangliu\}@pku.edu.cn}}

\newcommand{\yang}[1]{\textcolor{blue}{[Yang: #1]}}
\newcommand{\cam}[1]{\textcolor{blue}{#1}}

\maketitle

\begin{abstract}
Dynamic Scene Graph Generation (DSGG) focuses on identifying visual relationships within the spatial-temporal domain of videos. Conventional approaches often employ multi-stage pipelines, which typically consist of object detection, temporal association, and multi-relation classification. However, these methods exhibit inherent limitations due to the separation of multiple stages, and independent optimization of these sub-problems may yield sub-optimal solutions. To remedy these limitations, we propose a one-stage end-to-end framework, termed OED, which streamlines the DSGG pipeline. This framework reformulates the task as a set prediction problem and leverages pair-wise features to represent each subject-object pair within the scene graph. Moreover, another challenge of DSGG is capturing temporal dependencies, we introduce a Progressively Refined Module (PRM) for aggregating temporal context without the constraints of additional trackers or handcrafted trajectories, enabling end-to-end optimization of the network. Extensive experiments conducted on the Action Genome benchmark demonstrate the effectiveness of our design. The code and models are available at \url{https://github.com/guanw-pku/OED}.
\end{abstract}

\section{Introduction}
\label{sec:intro}
\begin{figure}[t]
    \includegraphics[width=\linewidth]{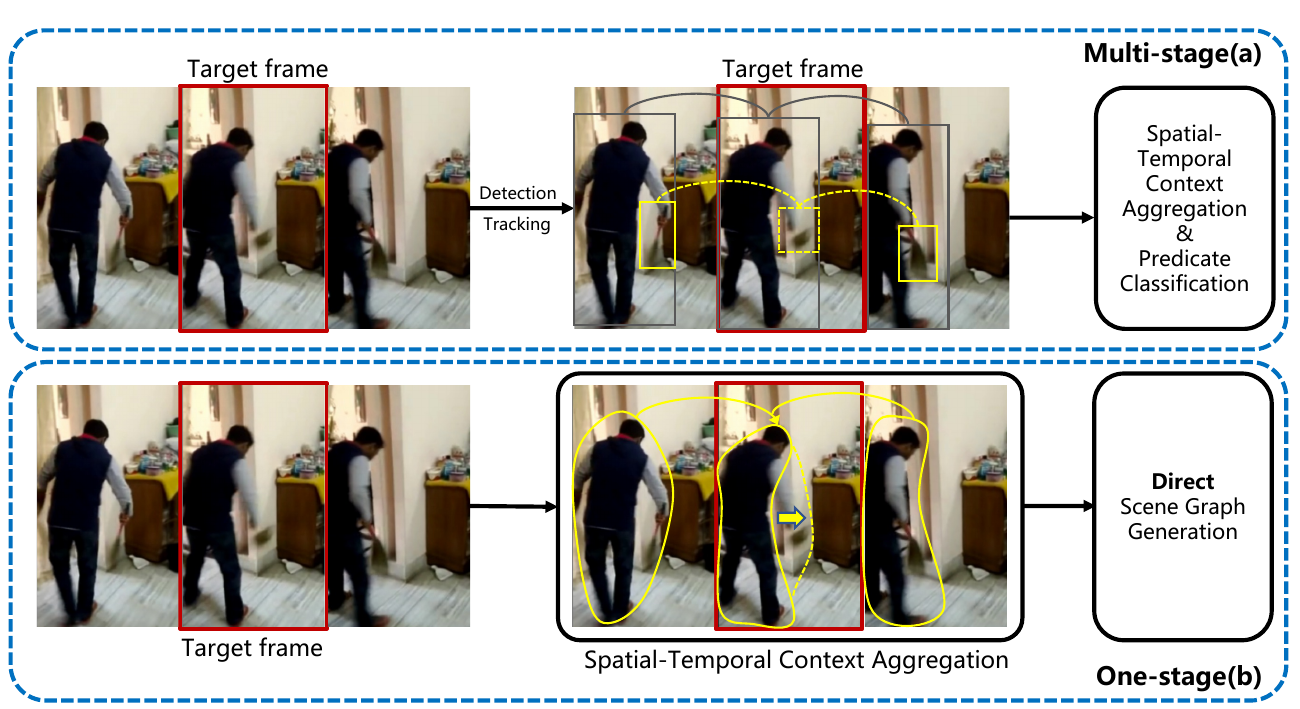}
    \caption{Comparison between existing multi-stage paradigm and proposed one-stage end-to-end framework. (a) Multi-stage methods, typically detect object instances by individual object detector and may associate objects between frames to aggregate temporal context based on detection results, followed by predicate classification for all candidate subject and object pairs, where tracking maybe lost. (b) Our one-stage end-to-end method, directly generates dynamic scene graph for given video sequence, without individual consideration for object instance detection and tracking. The missing spatial context and predicate temporal dependency could be supplemented with spatial context of reference frames.}
\label{fig:teaser}
\end{figure}

Scene Graph Generation (SGG) has emerged as a crucial component in advancing human-centric scene understanding, garnering significant research attention in recent years. The SGG research has been extensively employed in various high-level tasks, such as Visual Question Answering~\cite{lei2023symbolic, cherian20222, mao-etal-2022-dynamic, qian2022scene}, Visual Commonsense Reasoning~\cite{wang2022sgeitl} and Image Generation~\cite{farshad2023scenegenie}. Dynamic scene graph generation (DSGG) further extends SGG with additional temporal dimension and then becomes more challenging, which aims at understanding more informative spatial-temporal cues.

The primary objective of DSGG is to provide a sequence of comprehensive and structural representation of scenes by taking video sequence as input and detecting subject and object as nodes, as well as identifying the multi-relations between them as edges in graphs. Existing studies~\cite{li2022dynamic, feng2023exploiting} present promising results in DSGG  by decoupling this task into multiple stages: instance detection, temporal association, and multi-relation classification, as illustrated in Fig.~\ref{fig:teaser}(a). Specifically, subject and object detection results are obtained using an object detector. Subsequently, a temporal module, such as a tracker, establishes temporal links between instances in adjacent frames and aggregates temporal features on pair-wise combined subject-object proposals within the temporal sequence. The final stage entails performing multi-relation classification utilizing pair-wise features. Notably, prior to multi-relation classification, multi-stage methods requires the enumeration of instance or tracklet pairs. However, enumerative constructing all candidate subject-object pairs inevitably introduces not only a considerable of negative samples, significantly outnumbering positive ones, which is harmful in the training, but also substantial redundant computational costs. Furthermore, these methods suffer the problem of sub-optimal solutions due to independent optimization of separated multiple stages.

Recent research~\cite{zhang2023end} has proposed an end-to-end framework that unifies multiple tasks through a transformer-based structure. This approach first obtains instance results for each frame based on a transformer-based tracking model~\cite{zeng2022motr}. Subsequently, it enumerates subject-object pairs from tracked objects. Finally, the selected pairs from the previous frame are propagated to the target frame to aggregate temporal context and perform relation classification with pair-wise features. Nonetheless, this method still adopts a two-stage paradigm, constructing candidate subject-object pairs based on the detection results and subsequently executing relation classification. Such pairs are not always valid, rendering the pipeline more intricate and computationally expensive. 

Within the DSGG research community, the primary challenge lies in capturing temporal dependencies. Addressing this concern facilitates the detection of occluded or blurred objects and the perception of relation reliant on adjacent frames, such as \textit{looking at}, \textit{holding} and \textit{drinking from}. Existing methods have adopted complex yet sub-optimal strategies, including the utilization of trajectories or 3D convolution operators, to equip models with the capability to capture temporal dependencies. Nevertheless, trajectories generated by additional trackers are difficult for joint training, while 3D convolutions introduce substantial computational overhead, thereby limiting the overall efficiency and effectiveness of these approaches.

In this paper, we present a novel \textbf{O}ne-stage \textbf{E}nd-to-end architecture to directly predict \textbf{D}ynamic scene graphs through set prediction, termed \textbf{OED}, and introduce an effective temporal context aggregating strategy. OED reformulates dynamic scene graph generation as a set prediction problem by extending DETR~\cite{carion2020end} across both spatial and temporal dimensions. Our method comprises a Spatial Context Aggregation module and a Temporal Context Aggregation module, as shown in Fig.~\ref{fig:teaser}(b). The architecture first employs cascaded decoders to aggregate spatial context, with the former outputting pair-wise instance feature and the latter generating pair-wise relation feature. 
The pair-wise instance feature aggregates pair-wise instance related information and acts as the pair-wise relation query in Pair-wise Relation Decoder, providing a strong prior.
Subsequently, we concatenate both two features to obtain overall pair-wise feature and feed it into the proposed Progressively Refined Module~(PRM) for temporal context aggregating. 
PRM progressively selects pair-wise feature in reference frames and simultaneously optimizes the pair-wise feature in target frame to mine temporal dependencies via selected reference features, which implicitly links temporal information. 
This approach eliminates additional trackers and handcrafted trajectories, enabling end-to-end optimization of the network. 
Following this, classification heads and regression heads are utilized to predict DSGG results given spatial-temporal aggregated pair-wise feature. Finally, due to the challenge of incomplete annotations in video training data, we compute predicate classification loss only over a portion of the predictions that match ground truth, mitigating potential deterioration caused by missing annotations.

In summary, the primary contributions are as follows: (1) We introduce a simple one-stage end-to-end framework for DSGG, termed \textbf{OED}, which models dynamic scene graphs as a set prediction problem with pair-wise feature. (2) To effectively mine the temporal dependencies of relation, we propose a Progressively Refined Module~(PRM) for aggregating temporal context without the constraints of additional trackers, enabling end-to-end optimization of the network. (3) Experimental results on the Action Genome dataset demonstrate the superiority of the proposed one-stage end-to-end framework and the efficacy of the implemented temporal aggregation module.

\section{Related Work}
\subsection{Scene Graph Generation}
Scene Graph Generation for static image has caught broad attention since the benchmark was proposed~\cite{krishna2017visual}. 
Most previous works ~\cite{xu2017scene, yang2018graph, lin2020gps} adopt two-stage paradigms. 
The first stage is to detect all  objects by a off-the-shelf object detector, such as a pretrained Faster R-CNN~\cite{ren2015faster}.
Then the relationship between all candidate object pairs is classified using designed modules, such as Message Passing~\cite{xu2017scene, yang2018graph}, Graph Convolutional Network~\cite{lin2020gps}, Casual Inference~\cite{tang2020unbiased}.
Recently, some works~\cite{teng2022structured, cong2023reltr} adopts one-stage end-to-end paradigms to improve detection and predicate classification jointly without explicit detection.

However, additional temporal axis brings the need for effective perception and usage of complex spatial-temporal context, which means that SGG model hardly effectively handles DSGG task and thus the extension is not trivial.
\subsection{Dynamic Scene Graph Generation}
\subsubsection{Stacking Multi-stage Pipeline}
Dynamic scene graph generation task and its benchmark are proposed by~\cite{ji2020action}. Given that the context dependencies of this task span both spatial and temporal dimensions, the simultaneous modeling of spatial-temporal information and constructing spatial-temporal interaction relationships are essential for enhancing DSGG performance.

Previous approaches employed a multi-stage paradigm, utilizing object detectors to identify instances, and subsequently performing grouping and multi-relation classification on the detection results. 
STTran~\cite{cong2021spatial} leverages an off-the-shelf object detector to obtain instance detection results and then enhance and classify pairwise features via a spatial-temporal transformer. HOTR~\cite{ji2021detecting} introduces additional human pose features in the second stage to capture more relationship information. Some works~\cite{teng2021target, arnab2021unified} capture visual temporal dynamics from 3D CNN backbone, where TRACE~\cite{teng2021target} designs a hierarchical tree to aggregate spatial context and ~\cite{arnab2021unified} introduces message passing in a spatial-temporal graph to improve the spatial-temporal feature. TPT~\cite{zhang2023end} unifies object detection and object tracking together, thereby enhancing the pair-wise feature by the results of previous frame and aggregating rich spatial-temporal context. These methods enhance the instance features derived from object detection or achieve instance features via tracker directly, aggregate temporal information, and improve the accuracy of the classification. 

Nonetheless, they rely on dedicated modules for multiple tasks, thereby requiring individualized training schemes. This inevitably disrupts the collaborative interactions among these modules for different sub-tasks, ultimately resulting in sub-optimal solutions. On the other hand, the numerous interaction pairs generated by enumeration operation lead to substantial computational redundancy.

\subsubsection{Modeling Temporal Dependence}
In recent years, an increasing number of studies~\cite{li2022dynamic, wang2023cross,feng2023exploiting} have delved into the temporal information of features, constructing trajectory information to enhance inter-frame temporal dependencies. 
APT~\cite{li2022dynamic} proposes an anticipatory pre-training scheme to explore the temporal correlations among object pairs across different frames based on object tracking. $\text{TR}^2$\cite{wang2023cross} tracks object first and utilize the cross-modality feature from CLIP~\cite{radford2021learning} to guide the consistency between the temporal difference of pair-wise visual and semantic features. DSG-DETR ~\cite{feng2023exploiting} constructs inter-frame trajectories of object instances and pairs using bipartite graph matching, aiming to enhance the long-term temporal dependencies of temporal information and subsequently improve the performance of multi-relation classification. 

Nevertheless, trajectories generated by additional trackers are difficult for joint training, while handcrafted trajectories exhibit poor robustness, are prone to introducing noise, and consequently impact network performance. In this work, we formulate the DSGG as a one-stage set prediction task, utilizing pair-wise features to represent each subject-object pair within the scene graph. The one-stage framework eliminates the issue of inconsistent optimization objectives introduced by multi-stage approaches. Concurrently, we propose the PRM, which progressively filters reference frame pair-wise features and simultaneously optimizes target frame pair-wise features to mine temporal dependencies of relationships. By discarding additional trackers and handcrafted operators, notable performance enhancement is attained.

\section{Method}
\label{sec:method}

\begin{figure*}[!ht]
    \centering
    \includegraphics[width=\linewidth]{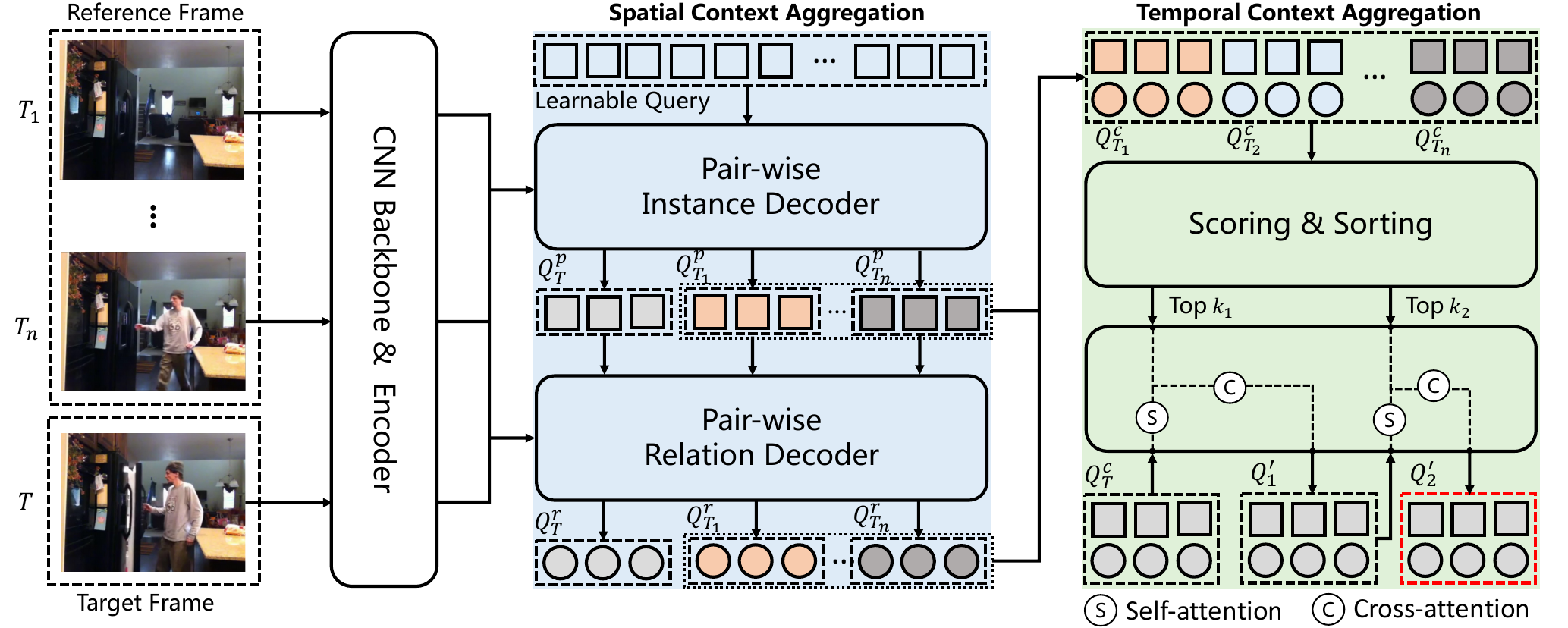}
    \caption{\textbf{OED Framework}: Spatial-temporal context aggregation is conducted within a one-stage end-to-end paradigm. Visual features of the target frame and reference frames are extracted using a CNN backbone and a Transformer encoder. Subsequently, two cascaded decoders are employed to aggregate spatial context both within and between pairs. Temporal context is then aggregated in a progressively refined manner, considering pair-wise features of the target frame and reference frames.}
\label{fig:pipeline}
\end{figure*}

\subsection{Problem Formulation}
\label{sec:formulation}
Dynamic scene graph generation aims to detect visual relations occurred in the target frame in video sequence.
Detected visual relations are represented by a special form of graph, called scene graph.
The nodes and edges in the scene graph refer to object instances and relations between them respectively, where an object instance consists of its label and spatial position. 
Therefore, a scene graph is equivalent to a list of triplets $\langle subject, predicate, object \rangle$, or $\langle s, p, o \rangle$ for short. 
To generate scene graphs, the target is to model the joint probability $P(\langle s, p, o\rangle|V)$ at each frame, where $\langle s, p, o\rangle$ belongs to a pre-defined triplet set and $V$ denotes the video sequence.

Some of previous works~\cite{cong2021spatial} factorize the joint probability as follows:
\begin{equation}
    P(\langle s, p, o\rangle|V) = P(p|s, o)P(s, o|D)P(D|V) 
\end{equation}
where $D$ represents object detection results in $V$. Recent works~\cite{feng2023exploiting, wang2023cross, zhang2023end} introduce additional tracking across frames to aggregate temporal information: 
\begin{equation}
    P(\langle s, p, o\rangle|V) = P(p|s, o)P(s, o|\mathcal{T})P(\mathcal{T}|D)P(D|V)
\end{equation}
where $\mathcal{T}$ represents object tracking results in $V$. 
These two types of solutions inevitably lead to multi-stage pipeline, which is sub-optimal due to separate training and upper bound of each stage.

In this work, we propose a one-stage method to directly model $P(\langle s, p, o\rangle|V)$. 
To utilize the temporal dependencies of predicate and alleviate the impact of motion and occlusion, we progressively aggregate temporal context information from reference frames. 
That is to say, we directly model $P(\langle s, p, o\rangle|I_i, \{I_{ref}\})$ at $i$-th frame, where $I_i$ indicates the $i$-th frame and $\{I_{ref}\}$ refers to the reference frames of $I_i$.

\subsection{Overview}
\label{sec:overview}
The pipeline of proposed approach is illustrated as Fig.~\ref{fig:pipeline}. 
Given the target frame and reference frames, OED directly generates scene graphs with spatial-temporal context in a way of set prediction.

First, the CNN backbone and Transformer encoder are sequentially utilized to extract visual features of each frame. 
To extract and aggregate useful spatial context, we adopt DETR-like~\cite{carion2020end} architecture and associate learnable queries with pair-wise feature of candidate object pairs. 
The pair-wise feature then extracts and aggregates spatial context in Transformer decoder.
To improve the detection of blurred object and predicate classification with dependencies on contextual frames at the same time, we introduce a progressive refined pair-wise feature interaction module~(PRM) to select and aggregate useful information from reference frames to the pair-wise feature of the target frame in a progressively refined way.
PRM fuses additional temporal context with the spatial aggregated pair-wise feature of the target frame, and then we obtain the final pair-wise feature with spatial-temporal context.

The pair-wise detection and predicate classification results will form a list of triplets $\langle s, p, o\rangle$, which corresponds to the scene graph of target frame.
\subsection{Spatial Context Aggregation}
\label{sec:spatial}
CNN visual backbone and Transformer encoder yield the visual features of each frame $\boldsymbol{F}=\{f_T, f_{T_1}, ..., f_{T_n}\}$, where $f_i\in \mathbb{R}^{HW\times d_{\text{model}}}, i\in \{T, T_1, ..., T_n\}$  and $(H, W)$ represents the scale of feature map.

In order to fully exploit the potential of set prediction design in the DSGG, we associate learnable queries $Q\in \mathbb{R}^{N_q\times d_\text{model}}$ with subject-object pairs $(s, o)$.
The pair-wise queries are to obtain specific visual features related to corresponding candidate pairs, which means spatial context aggregation.
In addition to aggregating the spatial information of each pair individually, the underlying connections between different pairs are significant as well, e.g. (\textit{person}, \textit{dish}) tends to co-occur with (\textit{person}, \textit{table}). 
We model and aggregate spatial context in these two ways using multi-head attention in transformer decoders. 

\textbf{Multi-head Attention.} Given query embedding $X_q$, key embedding $X_k$ and value embedding $X_v$, the output of multi-head attention is computed as follows:
\begin{equation}
\label{eq:mha}
\begin{gathered}
    \text{MHead}(X_q, X_k, X_v) = \text{Concat}(\text{head}_1,..., \text{head}_h)W \\
    \text{head}_i=\text{Attention}(X_qW_i^q, X_kW_i^k, X_vW_i^v) \\
    \text{Attention}(X_q, X_k, X_v) = \text{softmax} \big(\frac{X_qX_k^T}{\sqrt{d_k}}\big)X_v
\end{gathered}
\end{equation}
where $W_i^q\in \mathbb{R}^{d_{\text{model}}\times d_k}$, $W_i^k\in \mathbb{R}^{d_{\text{model}}\times d_k}$, $W_i^v\in \mathbb{R}^{d_{\text{model}}\times d_v}$ and $W\in \mathbb{R}^{hd_v\times d_{\text{model}}}$.

The spatial context dependencies between pairs are captured by multi-head self-attention $\text{SelfAttn}(Q)$ and Spatial Context Aggregation of each pair is performed by multi-head cross-attention $\text{CrossAttn}(Q, f_i)$ for $i$-th frame of given video sequence.
\begin{equation}
\label{eq:attn}
\begin{aligned}
    \text{SelfAttn}(Q)&=\text{MHead}(Q, Q, Q) \\
    \text{CrossAttn}(Q, f_i)&=\text{MHead}(Q, f_i, f_i) \\
\end{aligned}
\end{equation}



Considering that a single decoder struggles to handle two different tasks~\cite{zhang2021mining}, pair detection and predicate classification, we introduce two cascaded decoders. 
One is tailored to decode features for pair-wise instance related feature, while another one is for pair-wise predicate related feature.
Specifically, a set of learnable queries are used to capture pair-wise instance related information in Pair-wise Instance Decoder. 
Considering that pair-wise instance related feature can act a strong prior to predicate classification, we take it as pair-wise relation query to the Pair-wise Relation Decoder. 
In a word, the cascaded decoders aggregate the spatial context by pair-wise instance query and pair-wise relation query.

\textbf{Pair-wise Instance Decoder.} A set of learnable queries $Q$ extract and aggregate pair-wise instance related spatial context, as shown in Eq.~\ref{eq:attn}. 
Pair-wise instance feature from Pair-wise Instance Decoder $Q^p=\{\boldsymbol{q^p_1}, ..., \boldsymbol{q^p_{N_q}}\}\in \mathbb{R}^{N_q\times d_{\text{model}}}$ then acts as query of Pair-wise Relation Decoder. 

\textbf{Pair-wise Relation Decoder.} Apparently pair-wise instance could provide strong priors to classify predicate, especially for spatial and contacting predicates, such as (\textit{person}, \textit{chair}) with large overlapping area leading to \textit{sitting}. 
Thus, Pair-wise Relation Decoder take pair-wise instance feature $Q^p$ as query to capture and aggregate pair-wise relation specific spatial context $Q^r=\{\boldsymbol{q^r_1}, ..., \boldsymbol{q^r_{N_q}}\}\in \mathbb{R}^{N_q\times d_{\text{model}}}$, similar to the operations in the Pair-wise Instance Decoder.

Therefore, the spatial context information of triplets $\langle s, p, o\rangle$ corresponds to overall pair-wise feature $Q^c=\text{Concat}(Q^p,Q^r)=\{\boldsymbol{q^c_1}, ..., \boldsymbol{q^c_{N_q}}\}\in \mathbb{R}^{N_q\times2d_{\text{model}}}$, where $\boldsymbol{q^c_i}=\text{Concat}(\boldsymbol{q^p_i}, \boldsymbol{q^r_i}), i\in\{1, ..., N_q\}$.

\subsection{Temporal Context Aggregation}
\label{sec:temporal}
\begin{figure}[t]
    \includegraphics[width=\linewidth]{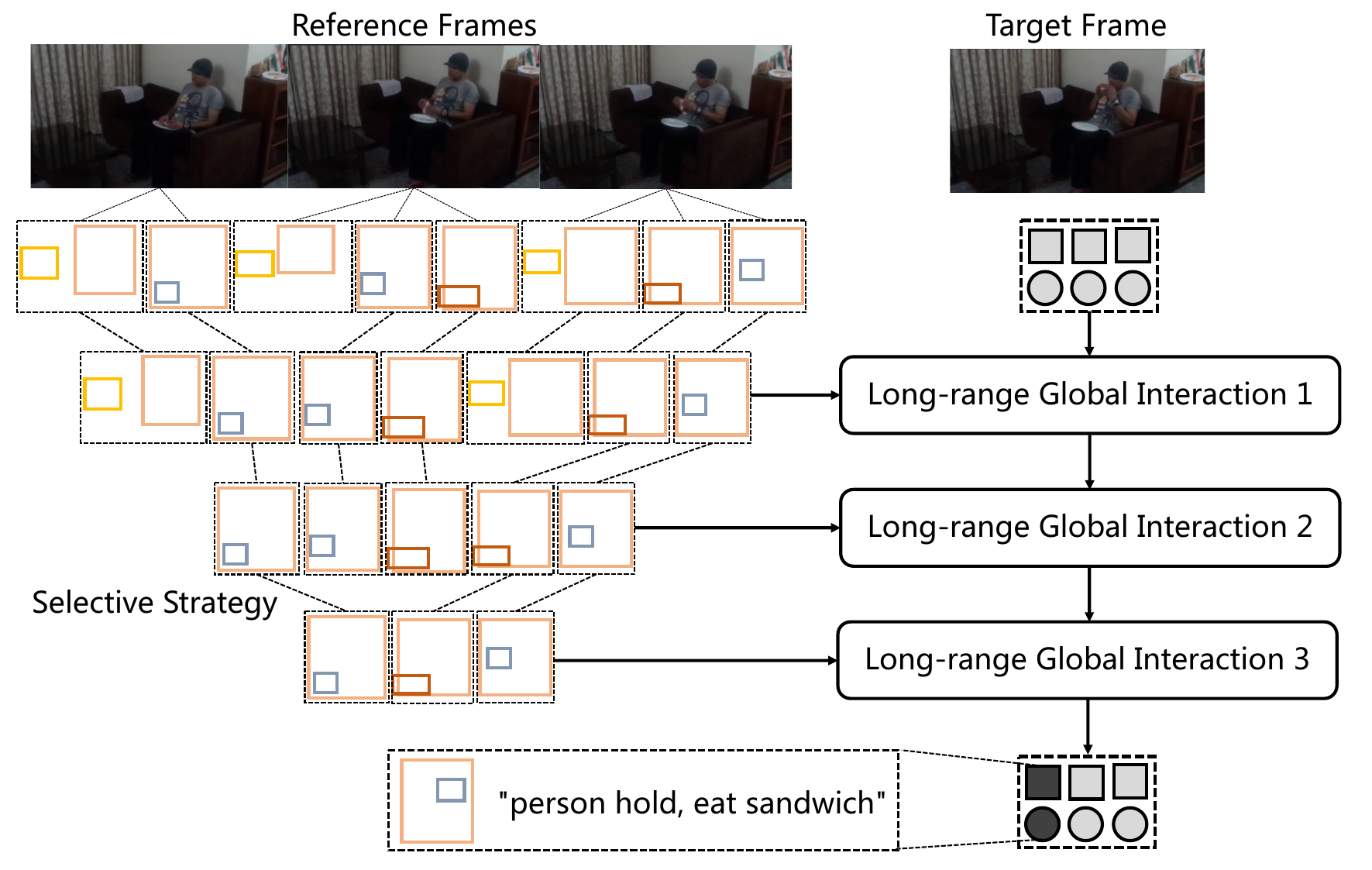}
    \caption{Progressively refined long-range global temporal context aggregation.}
\label{fig:prm}
\end{figure}
Through the cascaded decoders, the pair-wise feature $Q^c$ has aggregated rich spatial context information.
Besides spatial dependencies discussed in section~\ref{sec:spatial}, there are temporal dependencies of predicate across frames, e.g. (\textit{looking at - holding - drinking from}).
Reference frames could also improve the detection of blurred and occluded object in the target frame.
Therefore, this section further supplements pair-wise feature with additional temporal context, which is orthogonal to the spatial context.
To achieve this, we propose a multi-step progressively refined interaction module PRM, motivated by~\cite{deng2019relation}.


Specifically, we extract the spatial pair-wise feature of target frame and reference frames $\{Q^c_T, Q^c_{T_1}, ..., Q^c_{T_n}\}$ and concatenate the pair-wise features  of reference frames together $Q^c_{ref} = \{\boldsymbol{q^c_1}, ..., \boldsymbol{q^c_{n\times N_q}}\}$. Then we split the pair-wise feature $\boldsymbol{q^c_i}$ into pair-wise instance feature $\boldsymbol{q^p_i}$ and pair-wise relation feature $\boldsymbol{q^r_i}$ and use classification heads to score subject and object with $\boldsymbol{q^p_i}$ and score predicate with $\boldsymbol{q^r_i}$. 
We calculate the score of triplet by the multiplication of subject score $s_{sub}$, object score $s_{obj}$ and predicate score $s_{rel}$, and then rank them.
\begin{equation}
    \label{eq: triplet_score}
    p(\boldsymbol{q^c_i}) = s_{sub}\times s_{obj}\times s_{rel}
\end{equation} 
The pair-wise feature with higher score tends to have more correlations with corresponding ground truth.
Thus, we aggregate the temporal context from more confident reference pair-wise features.
Selecting a fixed number of reference pair-wise features is hard to hit a good balance and either bringing much noise or missing some informative reference pair-wise features, so we aggregate temporal context in a multi-step progressively refined way.

In $i$-th step, the selected Top-K reference pair-wise features $Q_{ref}^{c; k_i}$
interact with the pair-wise features in the target frame $Q_i^\prime$
in Transformer decoder progressively, which is formulated as
\begin{equation}
\begin{gathered}
    Q_i^\prime = \text{CrossAttn}(\text{SelfAttn}(Q_{i-1}^\prime), Q_{ref}^{c;k_i}), i > 0 \\
    Q_{ref}^{c;k_i}=\text{Top-K}(Q_{ref}^c, k_i) \\
    Q_0^\prime=Q_T^c
\end{gathered}
\end{equation}

The value of $k_i$ is gradually reduced to obtain more confident refined reference pair-wise features.
This progressively refined selection realizes the trade-off between more context information and less background noise. 

As shown in Fig.~\ref{fig:prm}, in the progressively refined process, some selected noises from reference frames, such as \textit{towel} denoted as yellow box, are gradually filtered out. 
Besides, PRM provides a way of long-range global temporal interaction, which means that the temporal interaction is not constrained inside the trajectories of object pair.
With the benefit of global perspective, PRM could capture the gradual movement of \textit{sandwich} from the \textit{dish} to \textit{person}.

After $m$ steps' progressively refined temporal aggregation, the pair-wise feature $Q_m^\prime$
is composed of abundant spatial-temporal context information. 
The spatial-temporal pair-wise feature is then divided into pair-wise instance feature $Q_T^p$
and pair-wise relation feature $Q_T^r$
, which are used to detect object pair and classify predicate respectively.

\subsection{Training and Inference}
\subsubsection{Training}
Given video sequence, OED generates a fxied set of predictions for each frame.
Then Hungarian algorithm finds out the optimal one-to-one matching $\hat{\sigma} \in \mathfrak{S}$ between prediction set $P=\{p_i\}_{i=1}^{N_q}$ and ground truth set $G=\{g_i\}_{i=1}^{N_q}$ that is padded with $\varnothing$ to the same length. 
\begin{equation}
    \hat{\sigma}=\underset{\sigma \in \mathfrak{S}_{N} }{\arg \min } 
 \sum_i^{N_q}\mathcal{L}_{\operatorname{match}}\left(g^i, p^{\sigma(i)}\right)
\end{equation}
where $\mathcal{L}_{\operatorname{match}}(g_i, p^{\sigma(i)})$ is the pair-wise cost between ground truth $g_i$ and the prediction with index $\sigma(i)$.
In the dynamic scene graph generation, we define the matching loss as:
\begin{equation}
    \mathcal{L}_{\operatorname{match}}\left(g^{i}, p^{\sigma(i)}\right)= \sum_{j \in \{\operatorname{s}, \operatorname{o}, \operatorname{p} \} }{\alpha_j \mathcal{L}_{\operatorname{cls}}^j} + \beta \sum_{j \in \{ \operatorname{s}, \operatorname{o} \}}{ \mathcal{L}_{\operatorname{box}}^{j}  }
\end{equation}
where 
$\mathcal{L}^j_{\operatorname{cls}}=\mathcal{L}^j_{\operatorname{cls}}\left(g^{i}_j, p_j^{\sigma(i)}\right), j\in \{s, o, p\}$ 
indicates the classification loss for subject, object and predicate, 
$\mathcal{L}^j_{\operatorname{box}}=\mathcal{L}^j_{\operatorname{box}}\left(g_j^{i}, p_j^{\sigma(i)}\right), j\in \{s, o, p\}$ 
indicates the bounding box regression loss of subject and object. 
We use cross entropy loss as classification loss of subject $\mathcal{L}_{\operatorname{cls}}^s$ and object $\mathcal{L}_{\operatorname{cls}}^o$, weighted sum of $L_1$ loss and GIoU loss~\cite{rezatofighi2019generalized} as bounding box regression loss of subject $\mathcal{L}_{\operatorname{box}}^{s}$ and object $\mathcal{L}_{\operatorname{box}}^{o}$ and focal loss~\cite{lin2017focal} as the classification loss of predicate. 

We adopt the matching loss as overall objective function but predicate loss.
Due to the incomplete annotation issue, we only calculate the predicate loss over the predictions matched with real ground truth, which is not padded with background.

\begin{equation}
    \mathcal{L}_{\operatorname{cls}}^{p'}\left(g^{i}, p^{\sigma(i)}\right) = \mathds{1}_{\{g^i\neq \varnothing\}} \mathcal{L}_{\operatorname{cls}}^{p}\left(g^{i}, p^{\sigma(i)}\right)
\end{equation}
The intuition here is that if the unmatched predictions are directly deemed as negative samples, the false negative samples keep misleading the model, which are positive samples at other frames instead.
The experiment in ablation studies shows that the matched predicate classification loss effectively mitigates the impact of incomplete annotation.

\subsubsection{Inference}
In the inference stage, there are a fixed number of pair predictions for each frame.
Because there may be multiple predicates for one pair, we rank the candidate triplets by scoring them as the multiplication of three-part confidences.
To reduce duplicate triplet detection, we filter out the predictions with lower scores that have the same triplet label and large overlapping area with others. 
Specifically, we take the multiplication of IoU of subject and object as the correlation in NMS to filter repeated predictions. 
The scene graph is generated by those retained triplet predictions.
\section{Experiments}
\begin{table*}[htbp]
\tiny
\centering
\caption{Comparison with state-of-the-art dynamic scene graph generation methods on Action Genome. The others methods follow multi-stage paradigms, while ours adopt a one-stage one. 
}
\setlength{\tabcolsep}{1pt}
\resizebox{0.8\textwidth}{!}{
\begin{tabular}{c cccccc cccccc}
\toprule
  & \multicolumn{6}{c}{With Constraint} & \multicolumn{6}{c}{No Constraints}\\
  \cmidrule(lr){2-7} \cmidrule(lr){8-13}
  Method  &
   \multicolumn{3}{c}{SGDET}  &  \multicolumn{3}{c}{PredCLS} & \multicolumn{3}{c}{SGDET}  & \multicolumn{3}{c}{PredCLS}\\ 
   \cmidrule(lr){2-4} \cmidrule(lr){5-7} \cmidrule(lr){8-10} \cmidrule(lr){11-13} 
   & R@10 & R@20 & R@50 & R@10 & R@20 & R@50  & R@10 & R@20 & R@50 & R@10 & R@20 & R@50 \\
\midrule
VRD~~\cite{lu2016visual}   &19.2 &24.5 &26.0  &51.7 &54.7 &54.7 &19.1 &28.8 &40.5 &59.6 &78.5 &99.2 \\
MSDN~\cite{li2017scene} &24.1 &32.4 &34.5  &65.5 &68.5 &68.5 &23.1 &34.7 &46.5 &74.9 &92.7 &99.0 \\
M-FREQ~\cite{zellers2018neural} &23.7 &31.4 &33.3 &62.4 &65.1 &65.1 &22.8 &34.3 &46.4 &73.4 &92.4 &\underline{99.6} \\
VCTree~\cite{tang2019learning} &24.4 &32.6 &34.7 &66.0 &69.3 &69.3 &23.9 &35.3 &46.8 &75.5 &92.9 &99.3 \\
RelDN~\cite{zhang2019graphical} &24.5 &32.8 &34.9 &66.3 &69.5 &69.5 &24.1 &35.4 &46.8 &75.7 &93.0 & 99.0 \\
GPS-Net~\cite{lin2020gps} &24.7 &33.1 &35.1 &66.8 &69.9 &69.9 &24.4 &35.7 &47.3 &76.0 &93.6 &99.5 \\
TRACE~\cite{teng2021target} &13.9 &14.5 &14.5 &27.5 &27.5 &27.5 &26.5 &35.6 &45.3 &72.6 &91.6 &96.4 \\
RelTR~\cite{cong2023reltr}  &19.7 &23.4 &25.9 &- &- &- &20.9 &24.6 &28.2 &- &- &- \\
STTran~\cite{cong2021spatial} &25.2 &34.1 &37.0 &68.6 &71.8 &71.8 &24.6 &36.2 &48.8 &77.9 &94.2 &99.1 \\
APT~\cite{li2022dynamic} &26.3 & \underline{36.1} & \underline{38.3} &69.4 &\underline{73.8} & \underline{73.8} &25.7 &37.9 &50.1 &78.5 &95.1 &99.2 \\
STTran-TPI~\cite{wang2022dynamic} &26.2 &34.6 &37.4 &69.7 &72.6 &72.6 &- &- &- &- &- &- \\
TR\textnormal{\^}2~\cite{wang2023cross} &26.8 &35.5 &\underline{38.3} &\underline{70.9} &\underline{73.8} &\underline{73.8} &27.8 &39.2 &50.0 &83.1 &\underline{96.6} &\bf{99.9} \\
TEMPURA~\cite{nag2023unbiased} &28.1 &33.4 &34.9 &68.8 &71.5 &71.5 &29.8 &38.1 &46.4 &80.4 &94.2 &99.4 \\
DSG-DETR~\cite{feng2023exploiting} &\underline{30.3} &34.8 &36.1 &- &- &- &\underline{32.1} &\underline{40.9} &48.3  &- &- &- \\
TPT~\cite{zhang2023end} &- &- &- &- &- &- &32.0 &39.6 &\underline{51.5} &\bf 85.6 &\bf 97.4 &\bf 99.9 \\
\midrule
Ours &\bf 33.5 &\bf 40.9 &\bf 48.9  &\bf 73.0 &\bf 76.1 &\bf 76.1 &\bf 35.3 &\bf 44.0 &\bf{51.8} &\underline{83.3} &95.3 &99.2 \\
\bottomrule
\end{tabular}
}
\label{tab:sota}
\end{table*}

\begin{table}[htbp!]
    \centering
    \setlength{\abovecaptionskip}{0.1 cm}
    \setlength{\belowcaptionskip}{-0.2 cm}
    \huge
    \resizebox{0.5\textwidth}{!}{
    \begin{tabular}{c|ccc|ccc|ccc}
    \toprule
    \multirow{2}{*}{\makecell[c]{\#}} &
    \multirow{2}{*}{\makecell[c]{Baseline}} &
    \multirow{2}{*}{\makecell[c]{SA}} &
    \multirow{2}{*}{\makecell[c]{TA}} & 
    \multicolumn{3}{c|}{With Constraint} & \multicolumn{3}{c}{No Constraints} \\
    &&&& R@10 &R@20 &R@50 &R@10 &R@20 &R@50 \\
    \midrule
    1 & \checkmark && &26.3 &29.2	&32.1 &28.4 &32.9	&37.2 \\
    2 & \checkmark & \checkmark & &31.5 &37.7 &43.7 &33.4 &41.6	&49.0 \\
    3 & \checkmark & \checkmark & \checkmark &33.5 &40.9 &48.9 &35.3 &44.0	&51.8 \\
    \bottomrule
    \end{tabular}}
    \caption{Ablation studies on our framework.}
    \label{tab:ablation_spatial_temporal}
\end{table}

\begin{table}[htbp!]
    \centering
    \setlength{\abovecaptionskip}{0.1 cm}
    \setlength{\belowcaptionskip}{-0.1 cm}
    
    \huge
    \resizebox{0.5\textwidth}{!}{
    \begin{tabular}{c|ccc|ccc|ccc}
    \toprule
    \multirow{2}{*}{\makecell[c]{\#}} &
    \multirow{2}{*}{\makecell[c]{Baseline}} &
    \multirow{2}{*}{\makecell[c]{CD}} &
    \multirow{2}{*}{\makecell[c]{ML}} & 
    \multicolumn{3}{c|}{With Constraint} & \multicolumn{3}{c}{No Constraints} \\
    &&&& R@10 &R@20 &R@50 &R@10 &R@20 &R@50 \\
    \midrule
    1 & \checkmark && &26.3 &29.2 &32.1 &28.4 &32.9	&37.2 \\
    2 & \checkmark & \checkmark & &27.2 &30.3 &33.7 &29.2 &34.0	&38.4 \\
    3 & \checkmark & \checkmark & \checkmark &31.5 &37.7 &43.7 &33.4 &41.6	&49.0 \\
    \bottomrule
    \end{tabular}}
    \caption{Ablation studies on Spatial Context Aggregation.}
    \label{tab:ablation_spatial}
\end{table}
\begin{table}[htbp!]
    \centering
    \setlength{\abovecaptionskip}{0.1 cm}
    \setlength{\belowcaptionskip}{-0.1 cm}
    \huge
    \resizebox{0.5\textwidth}{!}{
    \begin{tabular}{c|ccc|ccc|ccc}
    \toprule
    \multirow{2}{*}{\makecell[c]{\#}} &
    \multirow{2}{*}{\makecell[c]{SA}} &
    \multirow{2}{*}{\makecell[c]{NPR}} &
    \multirow{2}{*}{\makecell[c]{PR}} & 
    \multicolumn{3}{c|}{With Constraint} & \multicolumn{3}{c}{No Constraints} \\
    &&&& R@10 &R@20 &R@50 &R@10 &R@20 &R@50 \\
    \midrule
    1 & \checkmark && &31.5 &37.7 &43.7 &33.4 &41.6	&49.0 \\
    2 & \checkmark & \checkmark & &32.3	&39.7 &47.8 &34.0 &42.7 &50.6 \\
    3 & \checkmark & \checkmark & \checkmark &33.5 &40.9 &48.9 &35.3 &44.0	&51.8 \\
    \bottomrule
    \end{tabular}}
    \caption{Ablation studies on Temporal Context Aggregation.}
    \label{tab:ablation_temporal}
\end{table}
\begin{table}[htbp]
    \centering
    \setlength{\abovecaptionskip}{0.1 cm}
    \setlength{\belowcaptionskip}{-0.2 cm}
    \resizebox{0.45\textwidth}{!}{
    \begin{tabular}{c|ccc|ccc}
    \toprule
    \multirow{2}{*}{\makecell[c]{\#}} &
    \multicolumn{3}{c|}{With Constraint} & \multicolumn{3}{c}{No Constraints} \\
    &R@10 &R@20 &R@50 &R@10 &R@20 &R@50 \\
    \midrule
    1 &37.5 &44.8 &51.9 &40.1 &48.6 &56.5 \\
    2 &33.5 &40.9 &48.9 &35.3 &44.0 &51.8 \\
    \bottomrule
    \end{tabular}}
    \caption{Comparison between oracle query selection and progressively refined query selection.}
    \label{tab:oracle_select}
\end{table}

\subsection{Experimental Setting}
\textbf{Dataset:} We evaluate OED on the Action Genome (AG) dataset~\cite{ji2020action}, which annotates $234, 253$ frame scene graphs for sampled frames from around 10K videos, based on Charades dataset~\cite{sigurdsson2016hollywood}.
The annotations cover $35$ object categories and $25$ predicates.
The overall predicates consist of three types of predicates: attention, spatial and contacting.
There may be multiple spatial predicates or contacting predicates between the same pair.

\noindent\textbf{Evaluation Metrics:}
Following previous works~\cite{ji2020action,li2022dynamic,feng2023exploiting}, we adopt Recall@$k$ as evaluation metrics to measure the fraction of ground truth hit in the top $k$ predictions under \textit{With Constraint} and \textit{No Constraints} setting, where $k\in\{10, 20, 50\}$.
Specifically, We evaluate our method on two protocols: scene graph detection (SGDET) and predicate classification (PredCLS), following TPT~\cite{zhang2023end}.
SGDET aims to generate scene graphs for given videos, comprising detection results of subject-object pairs and the associated predicates. 
The localization of object prediction is considered accurate when the Intersection over Union~(IoU) between the prediction and ground truth is greater than 0.5. PredCLS, a simplified task to eliminate object detection errors,  requires methods to classify predicates for given oracle detection results of subject-object pairs.

\noindent\textbf{Implementation Details:}
We employ ResNet-50 as the CNN backbone. The Image Encode, Pair-wise Decoder and Relation Decoder consist of 6 transformer layers, with the number of predefined learnable query $N_q=100$. 
Following TPT~\cite{zhang2023end}, we initialize Image Encoder and Pair-wise Decoder with the weights pretrained on the MS-COCO dataset and subsequently fine-tune all modules on the Action Genome dataset. 
PRM includes three instances of progressively refind pair-wise interaction with Top-K as [$80n$, $50n$, $30n$] respectively, where $n$ denotes the number of reference frames. 
The threshold adopted in inference stage is $0.9$.
For the PredCLS task, aimed at predicting predicate labels for specified object pairs, we initialize the learnable queries using semantic features derived from the Glove embeddings of the given pair labels. Additionally, we incorporate position embeddings with spatial features obtained from the specified bounding boxes of the pairs. During inference, we derive the outputs by associating the labels and bounding boxes of the ground truth with the predicate classification results for the corresponding pairs.


\subsection{Comparison with State of the Arts}

We present a comparison of our results with state-of-the-art methods for dynamic scene graph generation in Tab.~\ref{tab:sota}. The performance in the SEDET task exemplifies the effectiveness of our approach. The SGDET task is aligned with the objective of dynamic scene graph generation, which entails generating scene graphs by aggregating spatial-temporal context from video sequences without incorporating additional information. In SGDET, our streamlined one-stage end-to-end pipeline surpasses other methods across all metrics under both \textit{With Constraint} and \textit{No Constraints} settings. OED outperforms the second-best methods by an average of $6.2\%$~($3.2\%$, $4.8\%$ and $10.6\%$ respectively) under the \textit{With Constraint} setting and an average of $2.2\%$ ($3.3\%$, $3.1\%$ and $0.3\%$ respectively) under \textit{No Constraints} setting. This outcome underscores the importance of addressing dynamic scene graph generation as a comprehensive task rather than partitioning it into multiple sub-tasks.
More performance comparison in SGDET task with long-tail issue related metircs can be found in supp.M section 1.

In PredCLS, OED improves the performance of the second-best methods by an average of $2.1\%$ ($2.0\%$, $2.2\%$ and $2.3\%$ respectively) under the \textit{With Constraint} setting. However, our method's performance in PredCLS is marginally lower than that of TPT and $\text{TR}^2$ under the \textit{No Constraints} setting. We conjecture the reason is as follows: in the PredCLS task, oracle object tracks from ground truth are provided. Both TPT and $\text{TR}^2$ are tracking-based methods, and they utilize the oracle trajectories in their respective track-based temporal aggregation modules. Due to the nature of the one-stage paradigm, our method cannot explicitly use this information, which consequently reduces the efficiency of leveraging oracle information. Furthermore, TPT employs additional multi-scale features and $\text{TR}^2$ incorporates the CLIP~\cite{radford2021learning} model, which is pre-trained using 4M image-text pairs, providing them with an advantage over our approach. Despite the fact that multi-stage methods benefit from oracle tracks and can directly aggregate entirely accurate spatial-temporal context, our approach still outperforms others by a significant margin under the \textit{With Constraint} setting and attains comparable performance under the \textit{No Constraints} setting.

\subsection{Ablation Study}

In this part, we evaluate the effectiveness of our designs in OED with SGDET task on Action Genome test set.

\noindent\textbf{Spatial-Temporal Context Aggregation:} In Tab.~\ref{tab:ablation_spatial_temporal}, We evaluate the effectiveness of the proposed Spatial Context Aggregation~(SA) and Temporal Context Aggregation~(TA) modules individually. We first adapt DETR~\cite{carion2020end} for dynamic scene graph generation, establishing it as our baseline~(\#1), where the object pair predictions and predicate classification are derived from the same decoded query representation. By incorporating the Spatial Context Aggregation module into the baseline~(\#2), we observe a significant improvement in performance. This indicates that the performance of spatial scene graph generation plays a crucial role in the effectiveness of dynamic scene graph generation. Furthermore, when the Temporal Context Aggregation module is integrated alongside the Spatial Context Aggregation module~(\#3), a further gain is achieved. This suggests that effectively exploiting temporal dependency information can further enhance the performance of DSGG.

\noindent\textbf{Designs in Spatial Context Aggregation:} In Tab.~\ref{tab:ablation_spatial}, we evaluate the efficacy of the proposed Cascaded Decoders~(CD) and Matched Predicate Loss~(ML). Building upon the baseline, we introduce an additional Pair-wise Relation Decoder and combine the two decoders in a cascaded manner~(\#2), addressing the optimization challenges of unified representation in multi-task settings and the dependence of predicate classification on pair detection results. The performance improvement achieved by the cascaded decoders validates our aforementioned considerations. 
More qualitative results can be found in supp.M section 2.
Furthermore, to mitigate the misleading effects of incomplete annotations in the Action Genome dataset, we compute the predicate loss only over the predictions from queries that match the ground truth~(\#3). The resulting performance gain underscores the effectiveness of the matched predicate classification loss in addressing the issue of incomplete annotations.


\noindent\textbf{Designs in Temporal Context Aggregation:} In Tab.~\ref{tab:ablation_temporal}, we evaluate the importance of temporal context and our proposed PRM. We select the spatial aggregation model as our baseline~(\#1). Taking into account the potential loss of information due to motion and the temporal dependency of predicates, we hypothesize that context clues can be obtained from adjacent reference frames to enhance pair detection and predicate classification. To incorporate the temporal context, we interact the pair-wise feature of target frame with all reference pair-wise features~(\#2) without progressively refined~(NPR). The experimental results substantiate the effectiveness of temporal context in dynamic scene graph generation. Moreover, considering that not all pair-wise features are valid, as they may attend to duplicate areas or background noise, we implement a progressively refined interaction~(PR) between the pair-wise features of target frame and reference frames~(\#3). The demonstrated effectiveness of the progressive refinement of pair-wise interactions indicates that filtering out background noise is crucial for improving semantic context aggregation.

\subsection{Discussion}

To further assess the effectiveness of temporal pair-wise interaction and estimate the upper bound of our PRM, we assume that the selected reference queries are oracle queries, meaning that these queries are matched with ground truth via bipartite matching. As illustrated in Tab.~\ref{tab:oracle_select}, the oracle selection~(\#1) achieves a significantly higher performance compared to our PRM~(\#2). This result indicates that there is still room for exploration and improvement in our approach. 
In the future, we plan to delve deeper into the effective selection of true positive samples from pair-wise features of reference frames. It is worth noting that our objective is not to obtain precise trajectories. We regard trajectories as a form of long-term yet local information, subject to instance-level constraints. It impedes the perception of the relationships among different instances across frames. We consider that long-term global information extracted by PRM plays a crucial role, and our approach focuses on filtering more accurate pair-wise instances across frames to facilitate the relation classification for the target frame.


\section{Conclusion}
In this paper, we present a one-stage end-to-end framework, named OED, for dynamic scene graph generation. Our approach reformulates the task as a set prediction problem and employs pair-wise features to represent each subject-object pair within the scene graph. Furthermore, we introduce a Progressively Refined Module~(PRM) for temporal context aggregating. The PRM progressively filters pair-wise features of reference frames while simultaneously optimizing the pair-wise features of the target frame to extract temporal dependencies through filtered features. Consequently, OED achieves significant improvement over the baseline, establishing sota performance across all metrics in SGDET task. 

\noindent \textbf{Acknowledgements.} This work was supported by the grants from the National Natural Science Foundation of China 62372014.

{\small
\bibliographystyle{ieee_fullname}
\bibliography{PaperForReview}
}


\end{document}